\documentclass[10pt,letterpaper]{article}

\usepackage[letterpaper,left=1.2in,right=1.2in,top=1in,bottom=1in,footskip=0.45in]{geometry}
\usepackage{iftex}
\ifPDFTeX
  \usepackage[T1]{fontenc}
  \usepackage[helvratio=0.9]{newtxtext}
\else
  \usepackage[helvratio=0.9]{newtxtext}
\fi
\usepackage{newtxmath}
\usepackage{amsmath}
\usepackage{graphicx}
\usepackage{booktabs}
\usepackage[table]{xcolor}
\usepackage{float}
\usepackage{array}
\usepackage{enumitem}
\usepackage{caption}
\usepackage{microtype}
\usepackage{titlesec}
\usepackage{fancyhdr}
\usepackage[skins]{tcolorbox}
\usepackage{tikz}
\usepackage[numbers,sort&compress,square]{natbib}
\usepackage[colorlinks=true,linkcolor=ohblue,citecolor=ohblue,urlcolor=ohblue]{hyperref}
\usepackage[capitalise,noabbrev,nameinlink]{cleveref}

\definecolor{ohink}{HTML}{1D1D1B}
\definecolor{ohmuted}{HTML}{6B6A66}
\definecolor{ohcap}{HTML}{55544F}
\definecolor{ohrule}{HTML}{BDBCB6}
\definecolor{ohpanel}{HTML}{FAFAF7}
\definecolor{ohpanelrule}{HTML}{E3E2DD}
\definecolor{ohgroup}{HTML}{F1F0EC}
\definecolor{ohblue}{HTML}{1F5FAD}
\definecolor{ohbluefill}{HTML}{EEF4FC}
\definecolor{ohteal}{HTML}{0F766E}

\titleformat{\section}{\sffamily\bfseries\large\color{ohink}}{\thesection}{0.75em}{}
\titleformat{\subsection}{\sffamily\bfseries\normalsize\color{ohink}}{\thesubsection}{0.75em}{}
\titleformat{\paragraph}[runin]{\normalfont\bfseries}{}{0pt}{}
\titlespacing*{\section}{0pt}{16pt plus 3pt minus 2pt}{7pt}
\titlespacing*{\subsection}{0pt}{11pt plus 2pt minus 2pt}{5pt}
\titlespacing*{\paragraph}{0pt}{5pt plus 1pt minus 1pt}{0.7em}
\setlist{itemsep=2pt,topsep=3pt,parsep=0pt,leftmargin=2.2em}

\arrayrulecolor{ohrule}
\fancypagestyle{firstpage}{%
  \fancyhf{}%
  \fancyfoot[L]{\sffamily\footnotesize{\color{ohteal}\rule{5pt}{5pt}}\hspace{5pt}{\color{ohmuted}An Openhagen technical report}}%
  \fancyfoot[R]{\small\color{ohmuted}\thepage}}

\newcommand{\base}{\mbox{Qwen3.5-9B}}
\newcommand{\ours}{\mbox{Concise-9B}}
\newcommand{\dci}[3]{\ensuremath{#1}~[\ensuremath{#2},\allowbreak\ \ensuremath{#3}]}
\newcommand{\x}{\ensuremath{\times}}
\newcolumntype{L}[1]{>{\raggedright\arraybackslash}p{#1}}
\newcommand{\grouprow}[2]{\rowcolor{ohgroup}\multicolumn{#1}{@{}l}{\textsc{\footnotesize #2}}\\}
\newcommand{\capt}[1]{\textbf{\color{ohink}#1}}
\newcommand{\hd}[1]{\textbf{\sffamily #1}}

\newcommand{\devN}{498}
\newcommand{\devGPQA}{198}
\newcommand{\devMATH}{150}
\newcommand{\devMMLU}{150}
\newcommand{\devRatio}{0.61}
\newcommand{\devBaseTok}{16.0K}
\newcommand{\devConTok}{9.8K}
\newcommand{\devSampleBase}{86.5}
\newcommand{\devSampleCon}{85.9}
\newcommand{\devSampleD}{\dci{-3.0}{-7.6}{+1.5}}

\newcommand{\devCescBmajD}{\dci{-2.0}{-9.0}{+4.5}}
\newcommand{\devCescBmajRatio}{0.36}

\newcommand{\devCescBsampleD}{\dci{+6.9}{+0.8}{+12.6}}
\newcommand{\devCescBsampleRatio}{1.46}

\newcommand{\devEscFiredBase}{11.3}
\newcommand{\devEscFiredCon}{10.6}

\newcommand{\devGPQATrail}{\dci{3.6}{0.5}{6.9}}
\newcommand{\devGPQAbase}{82.1}
\newcommand{\devGPQAcon}{80.3}
\newcommand{\devMATHD}{\dci{-0.4}{-3.1}{+2.2}}
\newcommand{\devMMLUD}{\dci{+1.0}{-0.5}{+2.6}}
\newcommand{\devFirstD}{\dci{-8.0}{-17.0}{+1.0}}

\newcommand{\heldUD}{\dci{-39}{-89}{+12}}
\newcommand{\heldUEq}{1}
\newcommand{\heldN}{8,983}
\newcommand{\heldD}{\dci{-18}{-69}{+32}}
\newcommand{\heldDpts}{\dci{-0.20}{-0.77}{+0.36}}
\newcommand{\heldRatio}{0.66}
\newcommand{\heldThinkRatio}{0.63}
\newcommand{\heldBaseAcc}{82.2}
\newcommand{\heldConAcc}{82.0}
\newcommand{\nHard}{1,276}
\newcommand{\dHard}{\dci{-4}{-28}{+20}}
\newcommand{\rHard}{0.69}
\newcommand{\eqHard}{2}

\newcommand{\trHardBase}{213}
\newcommand{\trHardCon}{93}
\newcommand{\duHard}{\dci{-25}{-49}{-1}}

\newcommand{\gainUHardBase}{\dci{+32}{+20}{+45}}
\newcommand{\gainUHardCon}{\dci{+11}{+5}{+18}}

\newcommand{\nConf}{2,275}
\newcommand{\dConf}{\dci{0}{-20}{+20}}
\newcommand{\rConf}{0.67}
\newcommand{\eqConf}{1}

\newcommand{\nMmlu}{5,432}
\newcommand{\dMmlu}{\dci{-14}{-54}{+26}}
\newcommand{\rMmlu}{0.64}
\newcommand{\eqMmlu}{1}

\newcommand{\trMmluBase}{214}
\newcommand{\trMmluCon}{32}

\newcommand{\nComp}{279}
\newcommand{\duComp}{\dci{-15}{-28}{-2}}

\newcommand{\nHardMath}{397}

\newcommand{\nSgpqa}{600}

\newcommand{\uHardLoss}{25}
\newcommand{\uHardTrail}{\dci{25}{1}{49}}
\newcommand{\uCompTrail}{\dci{15}{2}{28}}
\newcommand{\uStillBase}{21}
\newcommand{\uStillCon}{2}
\newcommand{\mmluInConf}{157}
\newcommand{\compCounts}{AIME 2024: 21, AIME 2025: 19, AIME 2026: 23, HMMT February 2025: 30, HMMT February 2026: 33, HMMT November 2025: 30, BRUMO 2025: 30, SMT 2025: 53 and CMIMC 2025: 40}
\newcommand{\uCompLoss}{15}
\newcommand{\rangeRatioLo}{0.61}
\newcommand{\rangeRatioHi}{0.80}
\newcommand{\ovHard}{3}
\newcommand{\ovConf}{35}
\newcommand{\ovMmlu}{427}
\newcommand{\unwrapCon}{8}
\newcommand{\unwrapBase}{0}
\newcommand{\fourSgpqaD}{\dci{-7}{-23}{+9}}
\newcommand{\fourSgpqaNinety}{\ensuremath{[-20,\ +6]}}
\newcommand{\fourSgpqaRatio}{0.61}
\newcommand{\fourHardN}{1,279}
\newcommand{\fourSgpqaRatioExact}{0.606}
\newcommand{\fourHardBaseStill}{128}
\newcommand{\fourSgpqaBaseStill}{9}

\newcommand{\fourHardTrail}{\dci{20}{1}{39}}

\newcommand{\fourBdevRatio}{0.57}
\newcommand{\fourBdevD}{\dci{-0.3}{-1.4}{+0.9}}
\newcommand{\fourBdevEq}{2}
\newcommand{\fourBmathRatio}{0.60}
\newcommand{\fourBsgN}{600}
\newcommand{\fourBsgRatio}{0.53}
\newcommand{\fourBsgD}{\dci{-1.3}{-3.2}{+0.6}}
\newcommand{\fourBsgMajD}{\dci{-2.7}{-5.7}{+0.3}}
\newcommand{\fourBdevTokBase}{18.2K}

\newcommand{\olmoSaveLo}{1}
\newcommand{\olmoSaveHi}{4}
\newcommand{\olmoDevTok}{9.8K}

\newcommand{\nineBdevTok}{16.0K}
\newcommand{\slackHardBase}{54}
\newcommand{\slackHardBaseCI}{[52, 56]}
\newcommand{\slackHardBaseCorr}{65}
\newcommand{\slackHardBaseWrong}{59}

\newcommand{\beforeHardBase}{88}

\newcommand{\slackHardCon}{41}

\newcommand{\slackConfBase}{64}
\newcommand{\slackConfBaseCI}{[62, 65]}
\newcommand{\slackConfBaseCorr}{67}
\newcommand{\slackConfBaseWrong}{59}

\newcommand{\beforeConfBase}{83}

\newcommand{\slackConfCon}{48}

\newcommand{\slackMmluCon}{48}
\newcommand{\medFracBase}{0.36}
\newcommand{\medFracCon}{0.56}
\newcommand{\medPctBase}{36}

\newcommand{\txtAfterLo}{93}
\newcommand{\txtAfterHi}{100}
\newcommand{\txtBeforeLo}{65}
\newcommand{\txtBeforeHi}{86}
\newcommand{\txtCovLo}{30}
\newcommand{\txtCovHi}{61}
\newcommand{\txtShareLo}{5}
\newcommand{\txtShareHi}{11}
\newcommand{\withinHard}{60}
\newcommand{\withinHardCI}{[54, 65]}
\newcommand{\withinHardPairs}{387}
\newcommand{\withinHardProblems}{196}
\newcommand{\withinDevBase}{54}
\newcommand{\withinDevBaseCI}{[48, 60]}

\newcommand{\withinDevCon}{60}
\newcommand{\withinDevConCI}{[54, 66]}

\newcommand{\oracleBaseLo}{0.35}
\newcommand{\oracleBaseHi}{0.44}
\newcommand{\oracleConLo}{0.49}
\newcommand{\oracleConHi}{0.56}
\newcommand{\kThreeLossLo}{1.5}
\newcommand{\kThreeLossHi}{9.2}
\newcommand{\kTwoLossLo}{3.4}
\newcommand{\kTwoLossHi}{13.7}
\newcommand{\kThreeHardBase}{\dci{-9.2}{-11.1}{-7.4}}
\newcommand{\kThreeBaseLossLo}{2.4}
\newcommand{\kThreeBaseLossHi}{9.2}
\newcommand{\kThreeHardBaseRatio}{0.45}
\newcommand{\conVsBaseHardRatio}{0.69}
\newcommand{\decShare}{86}
\newcommand{\decShareCI}{[83, 90]}
\newcommand{\decN}{3,551}
\newcommand{\decTb}{12,643}
\newcommand{\decTc}{8,229}
\newcommand{\decEb}{5,137}
\newcommand{\decEc}{4,538}
\newcommand{\decSb}{7,506}
\newcommand{\decSc}{3,692}
\newcommand{\decSlackCut}{51}
\newcommand{\decPreCut}{12}
\newcommand{\decShareHard}{81}

\newcommand{\decRobustHard}{90}
\newcommand{\decShareConf}{91}

\newcommand{\decRobustConf}{95}

\newcommand{\noReadoutBase}{402}
\newcommand{\noReadoutCon}{822}
\newcommand{\decTextSb}{1,176}
\newcommand{\decTextSc}{437}
\newcommand{\decTextSlackCut}{63}
\newcommand{\paceActual}{930}
\newcommand{\pacePred}{866}
\newcommand{\paceBaseActual}{934}
\newcommand{\paceRes}{\dci{+64}{+36}{+92}}

\newcommand{\paceNullLo}{+9}
\newcommand{\paceNullMid}{+17}
\newcommand{\paceNullHi}{+32}

\newcommand{\paceExcessCI}{\dci{+45}{+17}{+73}}
\newcommand{\paceLaneLo}{31}
\newcommand{\paceLaneHi}{54}

\newcommand{\budHardEight}{\dci{+27}{+1}{+53}}
\newcommand{\budHardSixteen}{\dci{+25}{0}{+50}}

\newcommand{\budHardTwelve}{\dci{+28}{+3}{+53}}
\newcommand{\budConfEight}{\dci{+22}{-1}{+46}}
\newcommand{\budConfSixteen}{\dci{+21}{0}{+42}}

\newcommand{\budConfTwelve}{\dci{+23}{0}{+46}}
\newcommand{\budEdgeMax}{9}
\newcommand{\saConfD}{\ensuremath{-5}}
\newcommand{\saConfCI}{\ensuremath{[-12,\ +2]}}
\newcommand{\saConfRatio}{0.67}

\newcommand{\saConfReadouts}{4,855}
\newcommand{\saConfCorrect}{1,979}
\newcommand{\saConfComp}{1,984}
\newcommand{\saConfCommit}{82.7}
\newcommand{\saConfVsSingle}{\dci{+12}{0}{+24}}
\newcommand{\saConfVsMaj}{\dci{-4}{-12}{+4}}
\newcommand{\saConfVsMajRatio}{0.52}
\newcommand{\saBaseD}{\ensuremath{-10}}
\newcommand{\saBaseCI}{\ensuremath{[-21,\ +1]}}
\newcommand{\saBaseRatio}{0.57}

\newcommand{\poolRows}{8,252}
\newcommand{\poolUnique}{7,752}
\newcommand{\roundProblems}{3,000}
\newcommand{\roundOneTargets}{4,230}
\newcommand{\roundTwoTargets}{4,690}
\newcommand{\roundHours}{1}
\newcommand{\roundDistinctOne}{2,945}
\newcommand{\roundDistinctTwo}{2,934}
\newcommand{\synthShare}{9}
\newcommand{\rdN}{39,546}
\newcommand{\rdAgree}{39,546}
\newcommand{\synN}{8,983}
\newcommand{\synOK}{8,833}
\newcommand{\synMiss}{150}
\newcommand{\synMissSuper}{146}
\newcommand{\nullReadouts}{30,476}
\newcommand{\fourBmathOv}{16}
\newcommand{\fourBsgOv}{0}

\newcommand{\ohmark}{%
  \begin{tikzpicture}[baseline=-0.6ex]
    \fill[ohteal] (0,0) circle (0.36cm);
    \begin{scope}[shift={(-0.21cm,-0.16cm)},x=0.013125cm,y=0.013125cm]
      \fill[white] (0,16) arc[start angle=180,end angle=360,radius=16] -- (24,16)
                   arc[start angle=0,end angle=-180,radius=8] -- cycle;
      \fill[white] (8,16) -- (24,16) -- (16,24) -- (9.333,24) .. controls (8.597,24) and (8,23.403) .. (8,22.667) -- cycle;
    \end{scope}
  \end{tikzpicture}}

\hypersetup{pdftitle={Concise-9B: Removing the Slack from a Reasoning Model's Thinking},pdfauthor={Alexander Boesgaard Lorup}}

\begin{document}
\thispagestyle{firstpage}

\begin{tcolorbox}[enhanced,colback=ohpanel,colframe=ohpanelrule,boxrule=0.5pt,arc=7pt,left=16pt,right=16pt,top=16pt,bottom=12pt]
{\sffamily\bfseries\fontsize{17}{21.5}\selectfont\color{ohink}Concise-9B: Removing the Slack\\ from a Reasoning Model's Thinking\par}
\vspace{9pt}
{\sffamily\bfseries\small\color{ohink}Alexander Boesgaard Lorup\par}
\vspace{2pt}
{\sffamily\small\color{ohmuted}Openhagen\par}
\vspace{11pt}
{\small
Reasoning models answer better when they think longer, but a long thinking trace keeps going well after its answer has formed.
We measure this slack in \base{} and remove most of it with concise self-training, two rounds of on-policy rejection-sampling fine-tuning that keep the shortest correct samples, plus a small share of answer-synthesis targets, which gives \ours{}.
On \heldN{} held-out problems (competition and hard mathematics, SuperGPQA, a five-area confirmation set and MMLU-Pro), \ours{} generates \heldRatio\x{} the tokens of \base{} at a 32K thinking budget, with accuracy equivalent within $\pm$1 point (\heldDpts{} points).
On a \devN-problem development suite it uses \devRatio\x{} the tokens, and single-sample accuracy, majority voting over three and four traces and pass@4 are all equivalent within $\pm$2 points (majority of two within $\pm$3).
To locate the slack, we force each trace on the hard and confirmation sets to answer every 2,048 thinking tokens and find the point after which each forced answer equals the final one.
For \base{}, \slackHardBase\% of thinking tokens on the hard set and \slackConfBase\% on the confirmation set come after that point; for \ours{}, \slackHardCon\% and \slackConfCon\%.
Stopping each trace there would keep every answer at \oracleBaseLo--\oracleBaseHi\x{} the base model's tokens, but finding that point takes hindsight: stopping as soon as three consecutive forced answers agree costs the base model \kThreeBaseLossLo--\kThreeBaseLossHi{} points.
Concise self-training removes slack without that cost: \decShare\% \decShareCI{} of the thinking tokens it saves come after the answer has formed.
It also has a limit: with traces continued to 80K tokens, \ours{} solves \uHardLoss{} fewer of \nHard{} hard problems (\duHard{}), mostly in competition mathematics, where long reasoning pays.
The same measurement gives a cheap commit signal for adaptive sampling: accepting a trace whose final answer agrees with its last two forced answers matched two-sample escalation within $\pm$1 point at \saConfRatio\x{} its tokens in a pre-registered test.
Results on Qwen3.5-4B, on which the recipe was developed, and a replication on Olmo-3-7B-Think support the recipe and suggest that the saving depends on how verbose the base model is.\par}
\vspace{9pt}
{\small\textbf{Date:} {\color{ohmuted}November 2026}\hfill\raisebox{-0.5ex}{\ohmark}\par}
\end{tcolorbox}
\vspace{4pt}

\section{Introduction}
\label{sec:intro}

Reasoning models trade tokens for accuracy.
A longer thinking budget lets a model keep reasoning where it would otherwise stop, often correcting itself, and this sequential test-time scaling is a large part of why thinking models are strong \citep{muennighoff2025s1,deepseek2025r1}; more generally, test-time compute spent well can let a model outperform a much larger one \citep{snell2024scaling}.
It is also expensive, and much of the expense may buy nothing: thinking models are known to keep reasoning on problems they have already solved \citep{chen2024overthinking,sui2025stopoverthinking}, and within a problem their shorter chains are more often right \citep{hassid2025overthink}.
Two questions follow.
How much of a thinking trace comes after its answer is already fixed?
And can a model be trained to drop that part without losing the reasoning that the rest of the trace does?

We answer both for \base{} \citep{qwen2026qwen35,qwen2026qwen359b}, an open-weight 9B thinking model, on whole evaluation sets: a \devN-problem development suite with eight samples per problem and model, and \heldN{} held-out problems.
\ours{} is \base{} after two rounds of \emph{concise self-training}: rejection-sampling fine-tuning \citep{zelikman2022star,yuan2023scaling,singh2023beyond}, run on-policy, that keeps the two shortest correct samples per problem, plus a small share of answer-synthesis targets, a form of shortest-correct selection \citep{munkhbat2025concise}.
The recipe was developed on Qwen3.5-4B \citep{qwen2026qwen35,qwen2026qwen354b} and applied to the 9B model unchanged, with the evaluated round fixed in advance.
To measure slack we use forced readouts: every 2,048 thinking tokens we close the trace's thinking and let the model state its answer, and we call the first point after which every readout gives the trace's final answer the point where the answer is \emph{fixed}.

\begin{figure}[tbp]
\centering
\includegraphics[width=\linewidth]{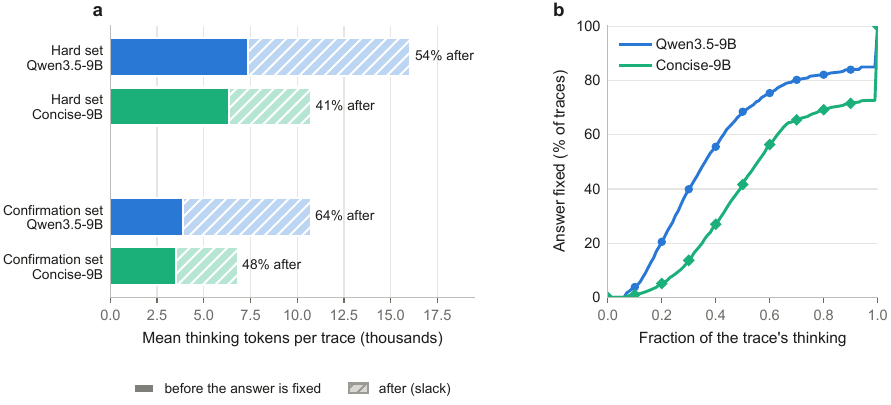}
\caption{\capt{Most thinking comes after the answer is fixed, and concise training removes mostly that part.}
(a)~Mean thinking tokens per trace at a 32K budget, split at the point after which every forced answer equals the trace's final answer; labels give the share after it (\cref{tab:slack}).
(b)~Share of traces whose answer is fixed by a given fraction of their thinking (hard and confirmation sets, \decN{} traces per model). A trace whose answer only settles at its end counts at 1.0.}
\label{fig:slack}
\end{figure}

\paragraph{Findings.}
\begin{itemize}
  \item \textbf{Fewer tokens at matched accuracy.} At a 32K thinking budget \ours{} generates \heldRatio\x{} the tokens of \base{} on \heldN{} held-out problems, with accuracy equivalent within $\pm$1 point (\heldD{} problems). On the development suite it uses \devRatio\x{} the tokens per sample, and single-sample accuracy, majority voting over three and four traces and pass@4 \citep{chen2021codex} are equivalent within $\pm$2 points (majority of two within $\pm$3; \cref{sec:results}).
  \item \textbf{Most of a trace is slack.} For \base{}, \slackHardBase\% (hard set) and \slackConfBase\% (confirmation set) of thinking tokens come after the answer is fixed; for half of its traces the answer is fixed within the first \medPctBase\% of the thinking (\cref{fig:slack,sec:slack}).
  \item \textbf{Concise training removes mostly slack.} Of the thinking tokens \ours{} saves, \decShare\% \decShareCI{} come after the answer is fixed. The part before it shrinks by only \decPreCut\%, and it is not simply cut short: at the lengths where \ours{} stops, it is right on \paceRes{} more hard problems than the base model's own answers at those lengths, \paceExcessCI{} problems beyond what further base-model samples show (\cref{sec:what}).
  \item \textbf{Trimming at inference needs hindsight.} Stopping each trace where its answer is fixed would keep every answer at \oracleBaseLo--\oracleBaseHi\x{} the tokens of \base{}, but stopping when three consecutive readouts agree loses \kThreeLossLo--\kThreeLossHi{} points (two: \kTwoLossLo--\kTwoLossHi{}; \cref{sec:trim}).
  \item \textbf{A limit on long problems.} With truncated traces continued to 80K tokens, \ours{} solves \uHardLoss{} fewer of \nHard{} hard problems (\duHard{}), \uCompLoss{} of them in competition mathematics (\cref{sec:heldout}).
  \item \textbf{A cheap commit signal.} A trace whose final answer agrees with its own last two readouts needs no second sample; this rule matched two-sample escalation in two pre-registered tests at \saConfRatio\x{} and \saBaseRatio\x{} its tokens (\cref{sec:selfagree}).
\end{itemize}
We claim no new training algorithm or stopping rule: concise self-training and shortest-correct selection exist \citep{munkhbat2025concise,yang2025thinkingoptimal,kimi2025k15}, and stopping when consecutive readouts agree follows the answer-consistency rule of \citet{liu2025convergence}, which we apply on a coarser readout grid (every 2,048 thinking tokens with $k=3$ or $k=2$; they read out at sentence boundaries and default to $k=10$).
What we add is a careful measurement of what such training removes, on a strong native thinking model and on sets large enough to bound the differences tightly.

\section{Related work}
\label{sec:related}

\paragraph{Overthinking and concise reasoning.}
Thinking models spend many tokens on easy problems \citep{chen2024overthinking}; \citet{sui2025stopoverthinking} survey remedies.
\citet{hassid2025overthink} find that shorter chains are more often correct and vote over the first chains to finish.
Self-training on short correct samples shortens reasoning: \citet{munkhbat2025concise} fine-tune non-thinking models on best-of-$N$ shortest correct samples, \citet{yang2025thinkingoptimal} select the shortest correct response for self-improvement, and \citet{kimi2025k15} fine-tune on the shortest correct response out of eight samples per question to make short-reasoning models more token-efficient.
Reinforcement learning controls length through the reward \citep{aggarwal2025l1,arora2025efficient,hou2025thinkprune}, and budget-aware objectives train for good answers at token budgets sampled across the trace \citep{qi2025anytime}.
Our recipe is the self-training kind; our contribution is the measurement of what it removes.

\paragraph{When the answer forms.}
Budget forcing ends thinking at a budget by appending an end-of-thinking delimiter and, optionally, an answer cue \citep{muennighoff2025s1}; we apply the same intervention, with our own cue, to read a trace's answer at many points.
Several methods stop a single trace once its intermediate answers converge \citep{liu2025convergence} or its confidence in a trial answer is high \citep{yang2025deer}, and DeepConf ends low-confidence traces early while voting over parallel traces \citep{fu2025deepconf}.
We measure how much such stopping could save in hindsight, how much a simple online rule saves, and how this compares with training.

\paragraph{Self-training and adaptive sampling.}
Rejection-sampling fine-tuning includes STaR \citep{zelikman2022star}, RFT \citep{yuan2023scaling}, RAFT \citep{dong2023raft} and ReST\textsuperscript{EM} \citep{singh2023beyond}; STaR, RAFT and ReST\textsuperscript{EM} iterate it, in the manner of expert iteration, which was introduced with tree search as the expert \citep{anthony2017thinking}.
Self-consistency returns the majority answer of sampled chains \citep{wang2023selfconsistency}, and adaptive variants stop sampling once a clear majority emerges or a window of samples agrees \citep{aggarwal2023adaptive,li2024escape}.
Our self-agreement rule uses a trace's own earlier answers as the agreement signal.

\section{Setup}
\label{sec:setup}

\paragraph{Model.}
\base{} \citep{qwen2026qwen35,qwen2026qwen359b} reasons in a delimited thinking block before answering; its 32 layers are eight repeats of three Gated DeltaNet layers \citep{yang2025gateddeltanet} and one gated-attention layer \citep{qiu2025gatedattention}.
We use each model's recommended thinking-mode sampling (for the Qwen3.5 models temperature 1.0, top-$p$ 0.95, top-$k$ 20 and presence penalty 1.5; for Olmo-3-7B-Think temperature 0.6 and top-$p$ 0.95) and call each independent draw a \emph{trace}.
Each prompt is the problem followed by ``Reason carefully. Give one final answer in \textbackslash boxed\{\ldots\}.''

\paragraph{Concise self-training.}
The training pool has \poolRows{} verifiable problems (\poolUnique{} unique) from the MATH and GSM8K training sets, MMLU test questions outside the development suite and GPQA questions outside GPQA-Diamond \citep{hendrycks2021math,cobbe2021gsm8k,hendrycks2021mmlu,rein2023gpqa}; \cref{app:training} gives its composition.
Each round uses \roundProblems{} fresh pool entries (\roundDistinctOne{} and \roundDistinctTwo{} distinct problems, since the hard MATH subset repeats MATH identifiers) and has three steps.
\begin{enumerate}
  \item \textbf{Collect.} The current model draws four samples per problem with an 8K-token budget, and makes two \emph{synthesis passes}: calls that read the four samples and re-derive the answer.
  \item \textbf{Select.} Keep the two shortest samples that are correct and finished naturally. Keep the correct synthesis passes of problems whose samples disagreed or whose majority was wrong, plus enough of the other correct synthesis passes to make up 30\% of the synthesis targets.
  \item \textbf{Train.} Fine-tune all parameters of the text backbone for one epoch on the kept targets, with the loss on generated tokens only (AdamW \citep{loshchilov2019adamw}, peak learning rate $2\times10^{-6}$, batches of 32 sequences).
\end{enumerate}
The two rounds kept \roundOneTargets{} and \roundTwoTargets{} targets, \synthShare\% of them synthesis passes.
\ours{} is the second round.
The recipe, including the number of rounds, was developed on Qwen3.5-4B (\cref{sec:replications}) and frozen before the 9B run.

\paragraph{Evaluation sets.}
The \emph{development suite} has \devN{} problems: all \devGPQA{} GPQA-Diamond questions \citep{rein2023gpqa} and fixed subsets of \devMATH{} MATH and \devMMLU{} MMLU test problems \citep{hendrycks2021math,hendrycks2021mmlu}.
On it each model drew four traces per problem with a 64K budget, in each of two sampling seeds; the first trace is also read at 32K (a 32K prefix of a longer trace is a valid 32K-budget trace).
The held-out sets played no role in training or in the recipe, and each model drew one trace per problem with a 32K thinking budget:
\begin{itemize}
  \item a \emph{hard set} of \nHard{} problems: \nComp{} final-answer competition problems (AIME 2024--2026, HMMT February 2025--2026 and November 2025, BRUMO, SMT and CMIMC 2025, from the public MathArena \citep{balunovic2025matharena} and Hugging Face H4 releases), \nHardMath{} hard MATH test problems and \nSgpqa{} SuperGPQA questions \citep{superGPQA2025};
  \item a \emph{confirmation set} of \nConf{} problems in five areas: MATH test problems, science (TheoremQA \citep{chen2023theoremqa} and SuperGPQA calculation questions), logical constraints (LogiQA \citep{liu2020logiqa} and generated relation and case-analysis tasks with programmatically verified answers), MMLU-Pro questions and executable code tasks scored by running the program;
  \item an \emph{MMLU-Pro set} of \nMmlu{} questions: every second of the 12,032 MMLU-Pro test questions (6,016) \citep{wang2024mmlupro}, without those excluded for overlap with our training data (\cref{app:data}) and those that are also in the confirmation set.
\end{itemize}
We checked every held-out problem against the problems used in training and excluded the \ovHard, \ovConf{} and \ovMmlu{} that overlap them (\cref{app:data}).
Each problem is counted once: the \mmluInConf{} MMLU-Pro questions that also belong to the confirmation set are kept only there, so the three held-out sets hold \heldN{} distinct problems.
For the \emph{uncapped} comparison, every trace that reached the 32K budget was continued from its own prefix to 80K tokens; \uStillBase{} \base{} traces and \uStillCon{} \ours{} traces still reach 80K.

\paragraph{Forced answers and readouts.}
When a trace reaches its budget we close its thinking, append ``The final answer is \textbackslash boxed\{'' and decode the answer greedily, as in budget forcing \citep{muennighoff2025s1}.
A \emph{readout} applies the same cue to a prefix of a trace: on every native trace of the hard and confirmation sets, and on \ours{}'s MMLU-Pro traces, we took one readout every 2,048 thinking tokens, up to 1,024 tokens before the trace's end.
We took readouts on the same grid on three further 64K \base{} traces per hard-set problem (\nullReadouts{} readouts; \cref{sec:scaling,sec:what,sec:selfagree}) and on the first of the 16K \ours{} confirmation-set traces used in \cref{sec:selfagree} (\saConfReadouts{} readouts).
A readout reuses the cached prefix and costs about 48 generated tokens.

\paragraph{Scoring.}
One checker scores every model: it takes the content of the last \textbackslash boxed\{\ldots\} in the committed answer, maps option letters to option text and compares the result with the reference answer after LaTeX normalization; code tasks compare exact outputs.
Votes and agreement tests use canonical answer keys, under which ``B'', ``(B)'' and the option's text agree, as do 0.5 and 1/2.
Forced answers are checked exactly like natural ones: re-scoring from their text the \rdN{} stored readouts of the analyzed native traces reproduces all \rdAgree{} stored verdicts (\cref{app:data}). We apply no post-hoc rescue to either model.

\paragraph{Statistics.}
All comparisons are paired by problem, with 95\% percentile intervals from 10,000 bootstrap resamples of problems \citep{efron1979bootstrap}; development-suite results average each problem's outcome over the two seeds before resampling.
Differences are in problems unless marked as points.
For equivalence we use two one-sided tests \citep{schuirmann1987tost} at $\alpha=0.05$ on 90\% paired intervals from 20,000 resamples, with margins of $\pm$1, $\pm$2 and $\pm$3 points of the set's size, and report the smallest margin that passes.
We say one system is more accurate than another only when the 95\% interval of the difference excludes zero.

\section{\ours{} against \base{}}
\label{sec:results}

\subsection{Development suite}
\label{sec:dev}

\begin{table}[tbp]
\centering
\caption{\capt{Development suite (\devN{} problems, 64K budget per trace, two sampling seeds pooled).}
Accuracy in \%; tokens are mean generated tokens per problem, summed over the traces a system uses.
$\Delta$ is \ours{} minus \base{} in problems, with a 95\% paired interval; the last column gives the smallest margin, in points, within which the two are equivalent.
Escalation (2\,$\to$\,4) draws two traces and answers if both finish and agree; otherwise it draws two more and takes the majority of four.}
\label{tab:dev}
\footnotesize
\setlength{\tabcolsep}{4.5pt}
\begin{tabular*}{\linewidth}{@{\extracolsep{\fill}}lrrrrlrc@{}}
\toprule
 & \multicolumn{2}{c}{\hd{\base{}}} & \multicolumn{2}{c}{\hd{\ours{}}} & & & \\
\cmidrule(lr){2-3}\cmidrule(lr){4-5}
\hd{System} & Acc. & Tokens & Acc. & Tokens & $\Delta$ [95\% CI] & Ratio & Equiv. \\
\midrule
Single sample (mean of 8 draws) & 86.5 & 16.0K & 85.9 & 9.8K & \dci{-3.0}{-7.6}{+1.5} & 0.61\x{} & $\pm$2 \\
Single trace, 32K budget & 84.8 & 14.2K & 84.5 & 9.4K & \dci{-1.5}{-10.5}{+7.0} & 0.66\x{} & $\pm$2 \\
Majority of 2 & 86.4 & 31.4K & 87.0 & 19.4K & \dci{+3.0}{-5.5}{+11.5} & 0.62\x{} & $\pm$3 \\
Majority of 3 & 87.9 & 47.8K & 88.0 & 29.3K & \dci{+0.5}{-6.0}{+7.0} & 0.61\x{} & $\pm$2 \\
Majority of 4 & 88.3 & 63.8K & 87.9 & 39.1K & \dci{-2.0}{-9.5}{+5.0} & 0.61\x{} & $\pm$2 \\
Escalation (2 $\to$ 4) & 88.4 & 38.7K & 87.9 & 23.3K & \dci{-2.5}{-9.5}{+4.0} & 0.60\x{} & $\pm$2 \\
\midrule
Pass@4 & 91.9 & 63.8K & 92.0 & 39.1K & \dci{+0.5}{-5.5}{+6.5} & 0.61\x{} & $\pm$2 \\
\bottomrule
\end{tabular*}
\end{table}

\begin{figure}[tbp]
\centering
\includegraphics[width=0.56\linewidth]{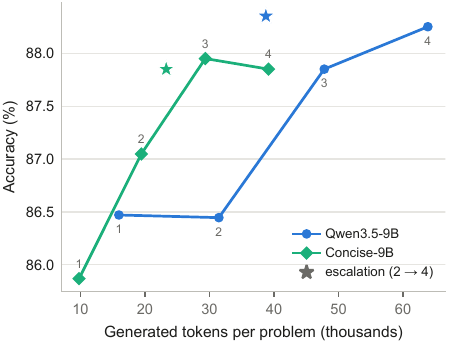}
\caption{\capt{Development suite: accuracy against generated tokens per problem.}
Lines are majorities of $k=1,\dots,4$ traces (numbers mark $k$; $k=1$ is the eight-draw single-sample mean); stars are escalation (2\,$\to$\,4). Values and paired intervals are in \cref{tab:dev}.}
\label{fig:frontier}
\end{figure}

\Cref{tab:dev} and \cref{fig:frontier} give the development-suite results.
\ours{} generates \devConTok{} tokens per trace against \devBaseTok{} for \base{} (\devRatio\x{}).
Its single-sample accuracy, averaged over eight draws per problem, is \devSampleCon\% against \devSampleBase\% (\devSampleD{} problems; the first draw alone at 64K, a noisier estimate, gives \devFirstD{}), and majority voting over two, three and four traces, escalation (2\,$\to$\,4) and pass@4 are all equivalent within $\pm$2 points, except the noisier majority of two ($\pm$3).
Escalation fires on \devEscFiredCon\% of problems for \ours{} and \devEscFiredBase\% for \base{}; \ours{} with escalation is equivalent within $\pm$2 points to the base model's majority of four (\devCescBmajD{}) at \devCescBmajRatio\x{} its tokens, and more accurate than the base model's single sample (\devCescBsampleD{}) at \devCescBsampleRatio\x{} its tokens.
The single-sample difference sits on GPQA-Diamond, where \ours{} trails by \devGPQATrail{} problems (\devGPQAcon\% against \devGPQAbase\% of \devGPQA{}); on MATH and MMLU it is \devMATHD{} and \devMMLUD{}.

\subsection{Held-out sets}
\label{sec:heldout}

\begin{table}[tbp]
\centering
\caption{\capt{Held-out sets: one trace per problem at a 32K thinking budget, and uncapped.}
Accuracy in \%; tokens are mean generated tokens per problem.
$\Delta$ is \ours{} minus \base{} in problems with a 95\% paired interval; ``Equiv.'' is the smallest equivalence margin in points.
``Trunc.'' counts traces that reached the budget (\base{} / \ours{}).
In the uncapped column every truncated trace is continued from its own prefix to 80K tokens. ``--'' in the Equiv. column means that no margin up to $\pm$3 points passes.}
\label{tab:heldout}
\footnotesize
\setlength{\tabcolsep}{3.2pt}
\begin{tabular*}{\linewidth}{@{\extracolsep{\fill}}lrrrrrlclc@{}}
\toprule
 & & \multicolumn{2}{c}{\hd{Accuracy}} & \multicolumn{2}{c}{\hd{Tokens}} & \multicolumn{3}{c}{\hd{32K budget}} & \hd{Uncapped} \\
\cmidrule(lr){3-4}\cmidrule(lr){5-6}\cmidrule(lr){7-9}\cmidrule(l){10-10}
\hd{Set} & $n$ & Qwen & Concise & Qwen & Concise & $\Delta$ [95\% CI] & Equiv. & Trunc. & $\Delta$ [95\% CI] \\
\midrule
\textbf{Hard set} & 1,276 & 73.2 & 72.9 & 16.9K & 11.7K & \dci{-4}{-28}{+20} & $\pm$2 & 213 / 93 & \dci{-25}{-49}{-1} \\
\quad Competition mathematics & 279 & 68.1 & 69.2 & 26.9K & 21.4K & \dci{+3}{-10}{+16} & -- & 124 / 70 & \dci{-15}{-28}{-2} \\
\quad Hard MATH & 397 & 98.2 & 97.0 & 11.8K & 7.6K & \dci{-5}{-10}{0} & $\pm$3 & 18 / 7 & \dci{-5}{-11}{+1} \\
\quad SuperGPQA & 600 & 59.0 & 58.7 & 15.6K & 9.8K & \dci{-2}{-21}{+18} & -- & 71 / 16 & \dci{-5}{-24}{+14} \\
\midrule
\textbf{Confirmation set} & 2,275 & 85.5 & 85.5 & 11.5K & 7.7K & \dci{0}{-20}{+20} & $\pm$1 & 117 / 26 & \dci{-2}{-22}{+18} \\
\quad MATH test & 476 & 98.1 & 98.5 & 12.7K & 7.7K & \dci{+2}{-5}{+9} & $\pm$2 & 16 / 0 & \dci{+2}{-5}{+9} \\
\quad Science & 600 & 66.3 & 66.0 & 16.9K & 11.3K & \dci{-2}{-16}{+12} & $\pm$3 & 80 / 20 & \dci{-3}{-17}{+11} \\
\quad Logical constraints & 575 & 87.8 & 87.5 & 8.4K & 6.5K & \dci{-2}{-11}{+7} & $\pm$2 & 4 / 4 & \dci{-2}{-10}{+6} \\
\quad MMLU-Pro (subset) & 289 & 83.4 & 84.1 & 11.4K & 7.2K & \dci{+2}{-8}{+12} & -- & 17 / 2 & \dci{+1}{-8}{+10} \\
\quad Executable code & 335 & 100.0 & 100.0 & 5.8K & 3.8K & \dci{0}{0}{0} & $\pm$1 & 0 / 0 & \dci{0}{0}{0} \\
\midrule
\textbf{MMLU-Pro} & 5,432 & 82.9 & 82.6 & 10.3K & 6.6K & \dci{-14}{-54}{+26} & $\pm$1 & 214 / 32 & \dci{-12}{-52}{+28} \\
\midrule
\textbf{All held-out} & 8,983 & 82.2 & 82.0 & 11.6K & 7.6K & \dci{-18}{-69}{+32} & $\pm$1 & 544 / 151 & \dci{-39}{-89}{+12} \\
\bottomrule
\end{tabular*}
\end{table}

\paragraph{At a 32K budget.}
Over all \heldN{} held-out problems \ours{} scores \heldConAcc\% against \heldBaseAcc\% (\heldD{} problems, equivalent within $\pm$1 point) at \heldRatio\x{} the generated tokens and \heldThinkRatio\x{} the thinking tokens (\cref{tab:heldout}).
Each set is equivalent on its own: within $\pm$\eqHard{} points on the hard set (\dHard{}) at \rHard\x{} the tokens, within $\pm$\eqConf{} on the confirmation set (\dConf{}) at \rConf\x{}, and within $\pm$\eqMmlu{} on MMLU-Pro (\dMmlu{}) at \rMmlu\x{}.
No area differs significantly; the token ratio ranges from \rangeRatioLo\x{} to \rangeRatioHi\x{} and is highest on competition mathematics, where both models often think to the budget.
\ours{} reaches the budget far less often: on \trHardCon{} hard problems against \trHardBase{}, and on \trMmluCon{} MMLU-Pro questions against \trMmluBase{}.

\paragraph{Continued to 80K.}
A 32K budget truncates the base model more often, so it can hide a difference in what each model reaches.
Continuing every truncated trace to 80K tokens changes little on the confirmation set and MMLU-Pro (\cref{tab:heldout}), and over all held-out problems uncapped accuracy is still equivalent within $\pm$\heldUEq{} point (\heldUD{}).
On the hard set \base{} gains \gainUHardBase{} problems from its \trHardBase{} truncated traces and \ours{} \gainUHardCon{} from its \trHardCon{}, so uncapped \ours{} trails by \uHardTrail{} problems.
The gap sits on competition mathematics (\duComp{} of \nComp{}), where long reasoning pays; on hard MATH and SuperGPQA the differences are small and not significant.
In a pre-registered comparison on the \nSgpqa{} SuperGPQA questions with four traces per model (first traces continued to 80K; three further 32K traces continued to 80K for \ours{}, three further 64K traces for \base{}), \ours{} with escalation is \fourSgpqaD{} problems from \base{} with escalation at \fourSgpqaRatio\x{} the tokens; the 90\% interval, \fourSgpqaNinety, does not establish equivalence within the registered $\pm$2 points, and the token ratio (\fourSgpqaRatioExact) misses the registered limit of 0.60.
Over the whole hard set (all \fourHardN{} problems, as registered) the same four-trace escalation trails by \fourHardTrail{} problems (\cref{app:tables}, \cref{tab:four}).
Concise training therefore gives up part of the base model's ceiling on the longest problems, a cost that a 32K comparison does not show.

\subsection{The development model and a replication}
\label{sec:replications}

\begin{table}[tbp]
\centering
\caption{\capt{The recipe on Qwen3.5-4B, where it was developed, and on Olmo-3-7B-Think.}
Accuracy in \%; $\Delta$ is the trained model minus its base model in points with a 95\% paired interval; ``Ratio'' is the ratio of generated tokens.
Development-suite rows pool two seeds and four traces per problem (Qwen3.5-4B, 64K budget) or two seeds (Olmo, 57,344-token budget); held-out MATH and SuperGPQA use one seed.
``Single sample'' is the mean over a problem's draws, ``single trace'' the first trace.
Olmo majorities vote over its three further traces, ties broken by trace order; Olmo intervals cluster the two seeds by problem.
Concise-4B is the model on which the recipe was developed, so its development-suite rows carry selection.}
\label{tab:rep}
\footnotesize
\begin{tabular*}{\linewidth}{@{\extracolsep{\fill}}lrlrrlr@{}}
\toprule
\hd{Set} & $n$ & \hd{System} & \hd{Base} & \hd{Trained} & $\Delta$ points [95\% CI] & \hd{Ratio} \\
\midrule
\grouprow{7}{Qwen3.5-4B $\to$ Concise-4B}
\quad Development suite & 498 & single sample & 84.2 & 84.0 & \dci{-0.3}{-1.4}{+0.9} & 0.57\x{} \\
\quad Development suite & 498 & majority of 4 & 86.8 & 86.7 & \dci{-0.1}{-1.6}{+1.4} & 0.57\x{} \\
\quad Held-out MATH & 484 & single sample & 98.1 & 97.7 & \dci{-0.4}{-1.1}{+0.3} & 0.60\x{} \\
\quad Held-out MATH & 484 & majority of 4 & 99.2 & 99.0 & \dci{-0.2}{-0.6}{0.0} & 0.60\x{} \\
\quad SuperGPQA & 600 & single sample & 54.0 & 52.7 & \dci{-1.3}{-3.2}{+0.6} & 0.53\x{} \\
\quad SuperGPQA & 600 & majority of 4 & 56.8 & 54.2 & \dci{-2.7}{-5.7}{+0.3} & 0.53\x{} \\
\grouprow{7}{Olmo-3-7B-Think $\to$ concise-trained}
\quad Development suite & 498 & single trace & 69.4 & 70.0 & \dci{+0.6}{-1.9}{+3.1} & 0.97\x{} \\
\quad Development suite & 498 & majority of 3 & 70.8 & 70.2 & \dci{-0.6}{-2.6}{+1.5} & 0.97\x{} \\
\quad SuperGPQA & 600 & single trace & 38.0 & 37.5 & \dci{-0.5}{-3.8}{+2.8} & 0.99\x{} \\
\quad SuperGPQA & 600 & majority of 3 & 40.2 & 40.7 & \dci{+0.5}{-2.7}{+3.7} & 0.96\x{} \\
\bottomrule
\end{tabular*}
\end{table}

\Cref{tab:rep} reports the recipe on two further models.
On Qwen3.5-4B, from which it was developed, Concise-4B uses \fourBdevRatio\x{} the tokens per trace on the development suite with single-sample accuracy equivalent within $\pm$\fourBdevEq{} points (\fourBdevD{} points), and \fourBmathRatio\x{} on held-out MATH.
On the \fourBsgN{} SuperGPQA questions its single sample is \fourBsgD{} points from the base model at \fourBsgRatio\x{} the tokens, and its majority of four \fourBsgMajD{}: not significant, but the same direction as the 9B four-trace comparison above.
On Olmo-3-7B-Think \citep{olmo2025olmo3}, a transformer from another developer with sliding-window and full attention and no linear-attention layers, every interval contains zero but tokens fall by only \olmoSaveLo--\olmoSaveHi\%.
Olmo is already terse, at \olmoDevTok{} tokens per development-suite trace against \fourBdevTokBase{} for Qwen3.5-4B and \nineBdevTok{} for \base{}.
Concise self-training can only select among a model's own correct samples, so it removes as much as the base model's spread in length allows: much of a verbose model's trace, little of a terse one's.

\section{Test-time slack}
\label{sec:slack}

\subsection{Longer budgets help; longer traces are not better}
\label{sec:scaling}

\begin{figure}[tbp]
\centering
\includegraphics[width=\linewidth]{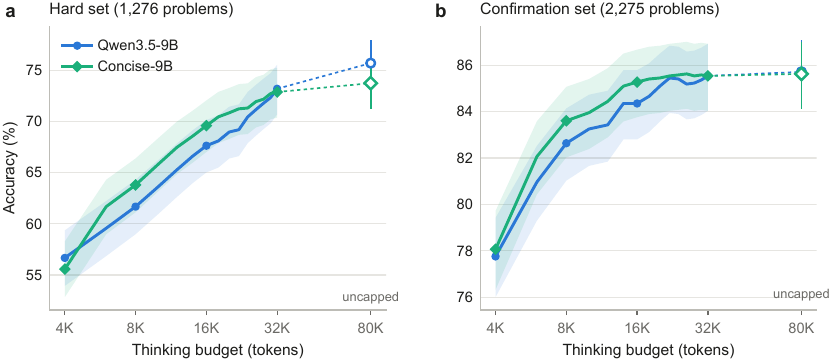}
\caption{\capt{Accuracy against thinking budget.}
A trace counts as correct at budget $B$ if it finished by $B$ with a correct answer, or if its latest readout at or before $B$ is correct; open markers continue truncated traces to 80K tokens. Bands are 95\% Wilson intervals \citep{wilson1927probable}; paired differences are in \cref{tab:budget}.}
\label{fig:budget}
\end{figure}

Readouts turn each trace into an accuracy-against-budget curve (\cref{fig:budget}).
Both models gain steadily at small budgets; on the hard set they keep gaining up to 32K tokens and beyond it, \base{} more than \ours{} (\cref{sec:heldout}), while on the confirmation set accuracy levels off: test-time scaling is real.
\ours{}'s curve lies above the base model's at intermediate budgets (\cref{tab:budget}): on the hard set the paired difference is \budHardEight{} problems at 8K, \budHardTwelve{} at 12K and \budHardSixteen{} at 16K; on the confirmation set it is \budConfEight{}, \budConfTwelve{} and \budConfSixteen{}. Only the hard-set intervals at 8K and 12K exclude zero.
At 32K the curves meet, and uncapped the base model is ahead on the hard set.

\begin{figure}[tbp]
\centering
\includegraphics[width=\linewidth]{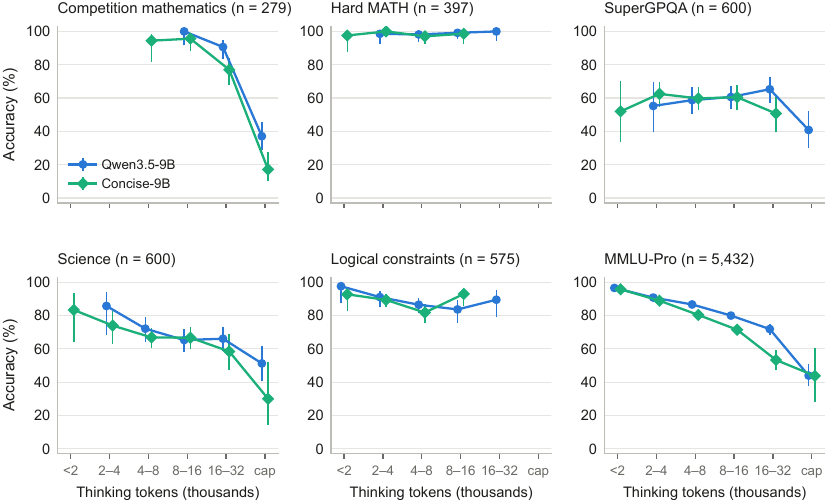}
\caption{\capt{Accuracy by thinking length, one trace per problem at a 32K budget.}
Bins hold traces that finished within the given number of thinking tokens; ``cap'' holds traces that reached the budget. Bins with fewer than 20 traces are not shown. Bars are 95\% Wilson intervals; counts are in \cref{tab:bins}.}
\label{fig:length}
\end{figure}

Across problems, however, longer traces are less often correct (\cref{fig:length}): on most benchmarks that are not saturated, accuracy falls with thinking length (on SuperGPQA it is flat or rising among traces that finish), and traces that reach the budget are the least accurate.
Part of this is difficulty, since hard problems draw long traces.
Within a problem the effect remains: of two traces of the same problem that disagree in correctness, the shorter one is correct in \withinHard\% \withinHardCI{} of \withinHardPairs{} such pairs (three 64K base-model traces per hard-set problem, \withinHardProblems{} problems with a discordant pair), and in \withinDevCon\% \withinDevConCI{} of pairs for \ours{} on the development suite (eight traces per problem; \withinDevBase\% \withinDevBaseCI{} for \base{}, an interval that includes 50\%).
A larger budget helps because it lets a long trace finish, not because length itself is productive, which suggests that much of a long trace does no work.

\subsection{Where the answer is fixed}
\label{sec:fixed}

\begin{table}[tbp]
\centering
\caption{\capt{Thinking before and after the answer is fixed (32K budget, one trace per problem).}
Mean thinking tokens per trace, split at the first readout from which every readout gives the trace's final answer (the end of the trace if there is none). ``After'' is the share of all thinking tokens after that point, with a 95\% interval; the next three columns give it for traces that finished correct, finished wrong, or reached the budget. ``Median'' is the median fraction of a trace's thinking at which its answer is fixed. MMLU-Pro readouts exist for \ours{} only.}
\label{tab:slack}
\footnotesize
\setlength{\tabcolsep}{3.4pt}
\begin{tabular*}{\linewidth}{@{\extracolsep{\fill}}llrrrlrrrr@{}}
\toprule
 & & \multicolumn{3}{c}{\hd{Thinking tokens}} & & \multicolumn{3}{c}{\hd{After, by outcome (\%)}} & \\
\cmidrule(lr){3-5}\cmidrule(lr){7-9}
\hd{Set} & \hd{Model} & All & Before & After & After (\%) [95\% CI] & Correct & Wrong & Cap & Median \\
\midrule
Hard set & Qwen3.5-9B & 16,057 & 7,379 & 8,677 & \dci{54.0}{51.9}{56.2} & 65 & 59 & 36 & 0.35 \\
Hard set & Concise-9B & 10,753 & 6,349 & 4,403 & \dci{41.0}{38.9}{43.1} & 46 & 41 & 28 & 0.55 \\
\addlinespace[2pt]
Confirmation set & Qwen3.5-9B & 10,729 & 3,880 & 6,849 & \dci{63.8}{62.4}{65.2} & 67 & 59 & 53 & 0.37 \\
Confirmation set & Concise-9B & 6,814 & 3,521 & 3,293 & \dci{48.3}{46.7}{49.9} & 50 & 44 & 44 & 0.56 \\
\addlinespace[2pt]
MMLU-Pro & Concise-9B & 5,916 & 3,065 & 2,851 & \dci{48.2}{47.1}{49.2} & 49 & 48 & 27 & 0.55 \\
\bottomrule
\end{tabular*}
\end{table}

\paragraph{Definition.}
Take a trace with final answer $a$ and readouts $r_1,\dots,r_k$ at $2{,}048, 4{,}096, \dots$ thinking tokens.
The answer is \emph{fixed} at the first readout $r_j$ such that $r_j,\dots,r_k$ all equal $a$; if $r_k \neq a$ or the trace has no readout, it is fixed at the end.
Two answers are equal if both are correct, or if both are wrong and share an answer key.
The thinking after this point is \emph{slack}: stopping there and forcing an answer would have returned $a$.
The 2,048-token grid makes the measure conservative, since an answer can be fixed up to one interval before the readout that first shows it.

\paragraph{Results.}
For \base{}, \slackHardBase\% \slackHardBaseCI{} of thinking tokens on the hard set and \slackConfBase\% \slackConfBaseCI{} on the confirmation set are slack (\cref{tab:slack}).
The answer is fixed before the end in \beforeHardBase\% and \beforeConfBase\% of traces, and for half of all traces, pooled over both sets, within the first \medPctBase\% of the thinking (\cref{fig:slack}b).
Slack is largest in traces that end correct (\slackHardBaseCorr\% and \slackConfBaseCorr\%), but wrong answers are also often fixed early (\slackHardBaseWrong\% and \slackConfBaseWrong\%): the model settles on an answer and then keeps checking it without changing it.
\ours{} has less slack, \slackHardCon\% and \slackConfCon\% (\slackMmluCon\% on MMLU-Pro), and its answers are fixed later in relative terms (pooled median \medFracCon{} of the trace against \medFracBase{}), because its traces end sooner after the answer forms.

\paragraph{Explicit statements come late.}
A text-based check gives a lower bound.
We located explicit answer statements in the thinking (``the answer is \ldots'', a boxed expression, ``option C is correct'') and took the first one after which every statement gives the final answer.
Such a statement precedes the end in only \txtCovLo--\txtCovHi\% of traces and marks only \txtShareLo--\txtShareHi\% of tokens as slack (\cref{tab:text}).
When it does occur, the readouts after it agree with the final answer \txtAfterLo--\txtAfterHi\% of the time, against \txtBeforeLo--\txtBeforeHi\% before it, so the statements are reliable; they are simply late.
The model would give its final answer long before it writes that answer down as its conclusion.

\subsection{Trimming slack at inference}
\label{sec:trim}

\begin{table}[tbp]
\centering
\caption{\capt{Stopping rules on single traces (32K budget).}
``Fixed'' stops each trace where its answer is fixed (hindsight; accuracy unchanged by construction). ``3 agree'' and ``2 agree'' stop at the first readout that repeats the previous two (one) readouts' answer key and return that readout (the answer-consistency rule of \citet{liu2025convergence}, here with $k=3$ or $k=2$ on the 2,048-token grid); traces where this never happens run to their end. Token ratios are against the same model's full traces and include 48 tokens per readout; $\Delta$ is in points against the full trace.}
\label{tab:stopping}
\footnotesize
\setlength{\tabcolsep}{3.6pt}
\begin{tabular*}{\linewidth}{@{\extracolsep{\fill}}llrrrlrlr@{}}
\toprule
 & & \multicolumn{2}{c}{\hd{Full trace}} & \hd{Fixed} & \multicolumn{2}{c}{\hd{3 agree}} & \multicolumn{2}{c}{\hd{2 agree}} \\
\cmidrule(lr){3-4}\cmidrule(lr){5-5}\cmidrule(lr){6-7}\cmidrule(l){8-9}
\hd{Set} & \hd{Model} & Acc. & Tokens & Ratio & $\Delta$ points [95\% CI] & Ratio & $\Delta$ points [95\% CI] & Ratio \\
\midrule
Hard set & Qwen3.5-9B & 73.2 & 16.9K & 0.44\x{} & \dci{-9.2}{-11.1}{-7.4} & 0.45\x{} & \dci{-13.7}{-15.9}{-11.5} & 0.30\x{} \\
Hard set & Concise-9B & 72.9 & 11.7K & 0.56\x{} & \dci{-6.7}{-8.3}{-5.3} & 0.62\x{} & \dci{-12.3}{-14.3}{-10.3} & 0.43\x{} \\
\addlinespace[2pt]
Confirmation set & Qwen3.5-9B & 85.5 & 11.5K & 0.35\x{} & \dci{-2.4}{-3.2}{-1.6} & 0.56\x{} & \dci{-4.5}{-5.6}{-3.5} & 0.40\x{} \\
Confirmation set & Concise-9B & 85.5 & 7.7K & 0.49\x{} & \dci{-1.5}{-2.2}{-0.8} & 0.75\x{} & \dci{-3.4}{-4.4}{-2.4} & 0.59\x{} \\
\bottomrule
\end{tabular*}
\end{table}

\begin{figure}[tbp]
\centering
\includegraphics[width=\linewidth]{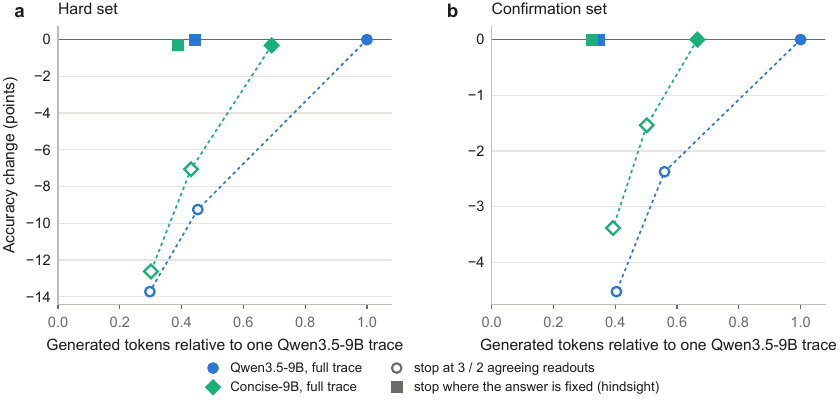}
\caption{\capt{Trimming by stopping rules and by training.}
Accuracy change and generated tokens relative to one full \base{} trace. Squares stop each trace where its answer is fixed, which needs hindsight; open markers stop when three or two consecutive readouts agree. \ours{}'s full trace (filled diamond) sits near the hindsight bound in accuracy without any stopping rule.}
\label{fig:stopping}
\end{figure}

If every trace stopped where its answer is fixed, accuracy would not change and \base{} would generate \oracleBaseLo--\oracleBaseHi\x{} its tokens (\ours{}: \oracleConLo--\oracleConHi\x{}; \cref{tab:stopping}).
This is an upper bound, on this readout grid, on what a rule that stops a single trace could save without changing any of its answers, because the stopping point is identified with knowledge of the trace's final answer.
A rule that sees only the readouts so far does much worse.
Stopping when three consecutive readouts agree \citep{liu2025convergence} changes \base{}'s accuracy on the hard set by \kThreeHardBase{} points at \kThreeHardBaseRatio\x{} its tokens, and stopping at two agreeing readouts loses more (\cref{tab:stopping}).
Early agreement is common on answers that the model later revises, so agreement alone does not tell a fixed answer from a provisional one.
Concise training trims differently (\cref{fig:stopping}): \ours{}'s full traces use \conVsBaseHardRatio\x{} the base model's tokens on the hard set at equivalent accuracy, without any stopping rule.

\subsection{What concise training removes}
\label{sec:what}

\begin{table}[tbp]
\centering
\caption{\capt{Where \ours{}'s token saving comes from (32K budget, thinking tokens per trace, paired by problem).}
$\Delta$ before and $\Delta$ after are \base{} minus \ours{} in the thinking before and after the answer is fixed, with 95\% intervals; ``Share after'' is the share of the total saving that comes after the answer is fixed. The last column repeats it on problems where both traces finished naturally and are at least 4,096 thinking tokens long ($n$ in parentheses).}
\label{tab:decomp}
\footnotesize
\setlength{\tabcolsep}{3.5pt}
\begin{tabular*}{\linewidth}{@{\extracolsep{\fill}}lrrrrlllr@{}}
\toprule
\hd{Set} & $n$ & \hd{Qwen} & \hd{Concise} & \hd{Saving} & $\Delta$ before [95\% CI] & $\Delta$ after [95\% CI] & Share after (\%) & Robustness \\
\midrule
Hard set & 1,276 & 16,057 & 10,753 & 5,304 & \dci{1{,}030}{686}{1{,}366} & \dci{4{,}274}{3{,}926}{4{,}624} & \dci{81}{75}{87} & 90 (694) \\
Confirmation set & 2,275 & 10,729 & 6,814 & 3,915 & \dci{358}{197}{522} & \dci{3{,}556}{3{,}347}{3{,}768} & \dci{91}{87}{95} & 95 (1,124) \\
Both & 3,551 & 12,643 & 8,229 & 4,414 & \dci{600}{439}{763} & \dci{3{,}814}{3{,}630}{3{,}999} & \dci{86}{83}{90} & 93 (1,818) \\
\bottomrule
\end{tabular*}
\end{table}

\paragraph{Mostly slack.}
\ours{} thinks \decTc{} tokens per trace against \decTb{} for \base{} on the hard and confirmation sets (\cref{tab:decomp}).
The thinking after the answer is fixed falls from \decSb{} to \decSc{} tokens (by \decSlackCut\%), the thinking before it from \decEb{} to \decEc{} (by \decPreCut\%), so \decShare\% \decShareCI{} of the saving is slack: \decShareHard\% on the hard set and \decShareConf\% on the confirmation set, and \decRobustHard\% and \decRobustConf\% on problems where both traces finished naturally and have at least 4,096 thinking tokens.
The grid counts a trace without readouts as having no slack, and \ours{} has more such traces (\noReadoutCon{} of \decN{}, against \noReadoutBase{} for \base{}); the grid therefore affects the two models differently, and we do not claim that the paired share is conservative.
The text-based measure, which sees only explicit statements, gives the same direction: the thinking after the first stable statement falls by \decTextSlackCut\%, from \decTextSb{} to \decTextSc{} tokens.

\paragraph{The rest is faster, not just shorter.}
The thinking before the answer is fixed also shrinks.
If \ours{} had simply learned to stop the base model's reasoning earlier, its accuracy would match the base model's own answers at the same lengths.
It does not.
At the length at which \ours{} stops on each hard-set problem, \base{}'s answer there (its readout, or its final answer if it had already finished) is correct on \pacePred{} problems, against \paceBaseActual{} at its own length; \ours{} is correct on \paceActual{}, \paceRes{} more than this prediction.
A natural conclusion beats a forced readout at the same length even for the same model, so we calibrate the comparison with three further 64K traces of \base{} per problem: they exceed the same prediction by \paceNullLo, \paceNullMid{} and \paceNullHi{} problems (\cref{tab:pace}).
\ours{} exceeds it by \paceExcessCI{} problems more than the average of these samples (\paceLaneLo--\paceLaneHi{} against each one): it reaches answers at lengths where the base model has not yet reached them.

\paragraph{Where reasoning was lost.}
Not all of the removed thinking was slack.
With traces continued to 80K, \ours{} trails by \uHardTrail{} problems on the hard set and by \uCompTrail{} on competition mathematics (\cref{sec:heldout}), where the base model's long traces often reach a correct answer after 32K tokens.
On the remaining held-out sets, uncapped accuracy is equivalent within $\pm$1 point.

\section{Using slack for adaptive sampling}
\label{sec:selfagree}

\begin{table}[tbp]
\centering
\caption{\capt{Self-agreement escalation, pre-registered tests.}
Correct problems and mean generated tokens per problem; $\Delta$ is against two-sample escalation with a 90\% paired interval (equivalence when it lies inside the stated margin), and ``Ratio'' its token ratio (registered limit 0.75). Readouts cost 48 tokens each. Both tests use all problems of their set, as registered, including the \ovConf{} confirmation-set and \ovHard{} hard-set problems excluded elsewhere for overlap with training; the registered evaluator uses 20,000 resamples for its intervals.}
\label{tab:selfagree}
\footnotesize
\begin{tabular*}{\linewidth}{@{\extracolsep{\fill}}lrrlr@{}}
\toprule
\hd{Policy} & \hd{Correct} & \hd{Tokens} & $\Delta$ [90\% CI] & \hd{Ratio} \\
\midrule
\grouprow{5}{Concise-9B, confirmation set (2,310 problems, 16K traces); margin $\pm$23 problems (1 point)}
\quad Single trace & 1,967 & 7.2K & -- & -- \\
\quad Majority of 3 & 1,983 & 21.6K & -- & -- \\
\quad Two-sample escalation (2 $\to$ 3), comparator & 1,984 & 16.6K & -- & -- \\
\quad {Self-agreement escalation, $m=1$} & 1,977 & 10.6K & \ensuremath{-7} \ensuremath{[-14,\ 0]} & 0.64\x{} \\
\quad \textbf{Self-agreement escalation, $m=2$} (primary) & 1,979 & 11.1K & \ensuremath{-5} \ensuremath{[-12,\ +2]} & 0.67\x{} \\
\quad {Self-agreement escalation, $m=3$} & 1,980 & 11.9K & \ensuremath{-4} \ensuremath{[-11,\ +3]} & 0.71\x{} \\
\grouprow{5}{Qwen3.5-9B, hard set (1,279 problems, 64K traces); margin $\pm$25 problems (2 points)}
\quad Single trace & 954 & 19.3K & -- & -- \\
\quad Majority of 3 & 984 & 57.6K & -- & -- \\
\quad Two-sample escalation (2 $\to$ 3), comparator & 985 & 44.3K & -- & -- \\
\quad {Self-agreement escalation, $m=1$} & 972 & 24.5K & \ensuremath{-13} \ensuremath{[-25,\ -1]} & 0.55\x{} \\
\quad \textbf{Self-agreement escalation, $m=2$} (primary) & 975 & 25.2K & \ensuremath{-10} \ensuremath{[-21,\ +1]} & 0.57\x{} \\
\quad {Self-agreement escalation, $m=3$} & 976 & 26.1K & \ensuremath{-9} \ensuremath{[-20,\ +2]} & 0.59\x{} \\
\bottomrule
\end{tabular*}
\end{table}

Adaptive sampling pays for its commit signal with extra samples \citep{aggarwal2023adaptive,li2024escape}. Our comparator here, \emph{two-sample escalation} (2\,$\to$\,3), answers when two traces finish and agree, and otherwise draws a third and takes the majority of three; it is the three-trace variant of the escalation in \cref{tab:dev}.
Because answers are usually fixed well before a trace ends, a trace can supply a second opinion on itself.
\emph{Self-agreement escalation} draws one trace and commits if it finished naturally and its readouts at the last $m$ checkpoints give its final answer; otherwise it falls back to two-sample escalation, drawing a second trace and, if the two disagree, a third and taking the majority (ties go to the shorter trace).
The readouts cost about 48 tokens each from the cached prefix.

We chose $m=2$ on the hard set with \ours{}, then registered the policy and its criteria before any readouts existed on the test data.
The primary test, on the confirmation set with \ours{} and 16K traces, required equivalence to two-sample escalation within $\pm$1 point at no more than 0.75\x{} its tokens.
It passed: self-agreement escalation scored \saConfCorrect{} problems against \saConfComp{} (\saConfD{}; 90\% interval \saConfCI{}) at \saConfRatio\x{} the tokens, committing on the first trace for \saConfCommit\% of problems (\cref{tab:selfagree}).
Against one trace the difference is \saConfVsSingle{} problems, and against a majority of three \saConfVsMaj{} at \saConfVsMajRatio\x{} its tokens.
A secondary test applied the same policy to \base{} on the hard set with 64K traces, with a $\pm$2-point margin; it also passed (\saBaseD{}; \saBaseCI{}) at \saBaseRatio\x{} the tokens.
The secondary test reuses the hard-set problems on which $m$ was chosen, with another model and new traces.
The saving is larger for the base model, consistent with its traces carrying more slack.

\section{Limitations}
\label{sec:limitations}

\paragraph{Models and data.}
The main results concern one base model; the recipe was developed on a smaller model of the same family and replicated on one model from another family, on fewer sets.
Each training round ran once, and our seeds are sampling seeds, not training seeds; the training targets include synthesis passes, which we did not ablate.
The development suite was used to develop the recipe on the 4B model, and its MATH and MMLU parts share source distributions with the training pool; the held-out sets did not influence the recipe.

\paragraph{Measuring slack.}
Readouts are greedy answers to a fixed cue at 2,048-token intervals, so the point where an answer is fixed is known only to that resolution, and traces shorter than 3,072 thinking tokens have no readout and count as having no slack.
Both choices make the measured slack smaller than the true one.
Slack is defined against the trace's own final answer, so it says nothing about whether that answer is right; for wrong answers, it measures checking that did not change the outcome.
The hindsight bound uses the full trace and is not a deployable rule.

\paragraph{Budgets and decoding.}
The main held-out comparisons use one trace per problem at a 32K budget, plus continuations to 80K; decoding is fixed to the model's recommended thinking-mode settings.
Token counts exclude prompt tokens, and we report tokens rather than wall-clock time.

\section{Conclusion}
\label{sec:conclusion}

Longer thinking makes \base{} more accurate, but most of a thinking trace comes after its answer is fixed: \slackHardBase--\slackConfBase\% of the base model's thinking tokens on the hard and confirmation sets.
Concise self-training removes mostly this part.
\ours{} generates about two thirds of the base model's tokens on \heldN{} held-out problems at accuracy equivalent within $\pm$1 point, and on the hard and confirmation sets \decShare\% of the thinking it drops came after the answer had formed; the reasoning before that point is shorter too, and reaches answers sooner than the base model's would at the same length.
Trimming the same slack at inference needs hindsight that simple stopping rules do not have, which makes training the more practical way to remove it.
The removal is not complete or free: on the longest competition problems, where the base model's extra thinking still finds answers, \ours{} gives up part of the base model's ceiling.
The measurement itself has a use: a trace that agrees with its own earlier answers needs no second sample, which makes adaptive sampling cheaper at equivalent accuracy.

\section*{Reproducibility}
Every number in this paper is computed by scripts from per-trace data files (correctness, token counts, truncation, readouts and the positions of answer statements for each trace, extracted from the stored evaluation records with the checker described in \cref{sec:setup}), except the Olmo results, the registered tests and the training statistics, which come from stored summaries of the evaluation records.
\Cref{app:training} lists the training data and hyperparameters, \cref{app:data} the evaluation sets, and \cref{app:stats} the statistical procedure.

\bibliographystyle{plainnat}
\bibliography{references}

\clearpage
\appendix
\crefalias{section}{appendix}

\section{Training details}
\label{app:training}

\begin{table}[H]
\centering
\caption{\capt{Training pool.} The hard MATH subset shares identifiers with MATH training problems, hence the smaller number of unique problems.}
\label{tab:pool}
\footnotesize
\begin{tabular}{@{}lr@{}}
\toprule
\hd{Source} & \hd{Problems} \\
\midrule
MATH training set & 3,002 \\
MATH training set, hard subset & 500 \\
GSM8K training set & 1,500 \\
MMLU test questions outside the development suite & 3,000 \\
GPQA main set, outside GPQA-Diamond & 250 \\
\midrule
Total (unique problems) & 8,252 (7,752) \\
\bottomrule
\end{tabular}
\end{table}

\begin{table}[H]
\centering
\caption{\capt{The two rounds of concise self-training on \base{}.} Each round draws \roundProblems{} pool entries (\roundDistinctOne{} and \roundDistinctTwo{} distinct problems, since the hard MATH subset repeats MATH identifiers), collects four 8K samples and two synthesis passes per entry and keeps the targets described in \cref{sec:setup}.}
\label{tab:rounds}
\footnotesize
\begin{tabular*}{\linewidth}{@{\extracolsep{\fill}}lrrrrrrr@{}}
\toprule
 & & \multicolumn{2}{c}{\hd{Collected}} & \multicolumn{2}{c}{\hd{Kept targets}} & & \\
\cmidrule(lr){3-4}\cmidrule(lr){5-6}
\hd{Round} & \hd{Entries} & Samples & Synthesis & Samples & Synthesis & \hd{Target tokens} & \hd{Steps} \\
\midrule
Round 1 & 3,000 & 12,000 & 6,000 & 3,833 & 397 & 14.7M & 133 \\
Round 2 (Concise-9B) & 3,000 & 12,000 & 6,000 & 4,273 & 417 & 15.1M & 147 \\
\bottomrule
\end{tabular*}
\end{table}

\paragraph{Optimization.}
Full-parameter supervised fine-tuning of the text backbone with token-summed cross-entropy on generated tokens, normalized by the batch's target tokens; AdamW ($\beta_1=0.9$, $\beta_2=0.95$, no weight decay) with a peak learning rate of $2\times10^{-6}$, linear warm-up and cosine decay to 10\% of the peak; batches of 32 sequences, one epoch per round, gradient-norm clipping at 1.0, bf16 mixed precision.
Each round's training took about \roundHours{} hour on one NVIDIA B200 GPU.

\paragraph{Synthesis passes.}
A synthesis pass has an 8K budget with thinking enabled. It sees the last part of each sample's reasoning and up to 400 characters of each sample's answer, and is instructed: ``Several independent attempts at this problem are shown above. They may disagree, and any of them may be wrong. Using them only as evidence, re-derive the answer yourself and check the decisive steps. Give one final answer in \textbackslash boxed\{\ldots\}.''
The evaluated systems use no synthesis passes.

\paragraph{Other models.}
Concise-4B and the Olmo model were trained with the same pool, collection budget, selection rule and hyperparameters, and the second round is the evaluated model in both cases. Olmo-3-7B-Think has a 64K context, so its largest evaluation budget is 57,344 generated tokens.

\section{Evaluation sets and decontamination}
\label{app:data}

The training pool excludes every problem that matches a development-suite problem by dataset identifier or by whitespace- and case-normalized text.
The hard set's competition problems were de-duplicated and 13-gram decontaminated against our training data, leaving \compCounts{} problems; its hard MATH problems exclude the development suite and any MATH problem in the training pool.
Its SuperGPQA questions were drawn at random from the dataset.
The confirmation set was built once, before any experiment on it, and is decontaminated by identifier and normalized text against the training pool, the development suite and the hard set; its generated tasks are re-derived by a verifier before use.
Its MMLU-Pro questions come from the same test split as the MMLU-Pro set, and \mmluInConf{} of the analyzed ones appear in both; we count them once, in the confirmation set.
Finally we compared every held-out problem with the problems actually used in the two training rounds by word 8-gram and prefix overlap, and excluded the \ovHard{} hard-set, \ovConf{} confirmation-set and \ovMmlu{} MMLU-Pro problems that overlap; most are MATH test problems that nearly duplicate training problems and MMLU-Pro questions derived from MMLU test questions.
The same check against the problems used in Concise-4B's two training rounds flags \fourBmathOv{} held-out MATH and \fourBsgOv{} SuperGPQA problems, which the 4B comparisons exclude.
On the \synN{} analyzed held-out problems, a synthetic forced answer that states the reference answer is accepted on \synOK{}; \synMissSuper{} of the \synMiss{} exceptions are SuperGPQA questions whose reference option text is malformed LaTeX, on which the checker still credits the correct option letter. The checker rejects \unwrapCon{} of \ours{}'s held-out answers and \unwrapBase{} of \base{}'s that wrap a correct option in \textbackslash text\{\} inside the box; we leave these as scored.

\section{Statistical methodology}
\label{app:stats}

\paragraph{Paired bootstrap.}
For two systems evaluated on the same problems we resample problems with replacement and sum the per-problem differences, 10,000 times for 95\% intervals and 20,000 times for the 90\% intervals of equivalence tests; intervals are the percentiles.
The resamples of a comparison are seeded by its data, so a comparison that appears in two tables has the same interval in both.
The two registered evaluators (\cref{tab:selfagree,tab:four}) use 20,000 resamples for all their intervals.
On the development suite each problem's outcome is first averaged over the two sampling seeds (and, for single-sample accuracy, over the four traces of each seed).
Ratios of sums, such as the share of slack, use the same resampling.
Single-model accuracies in the figures carry 95\% Wilson intervals \citep{wilson1927probable}.

\paragraph{Equivalence.}
Two systems are equivalent within $\pm m$ points when the whole 90\% interval of their difference lies strictly inside $\pm m\%$ of the set's size, which is a two one-sided test at $\alpha=0.05$.

\paragraph{Votes.}
Majority votes count traces with an answer key, ties go to the shorter trace, and a problem with no keyed trace counts as wrong.
Escalation commits only on traces that finished naturally; a forced answer never triggers a commitment.

\paragraph{Budget curves.}
A trace still thinking at budget $B$ is scored by its latest readout at or before $B$.
Readouts stop 1,024 tokens before a trace's end, so a trace that ends within that stretch after $B$ is scored by its previous readout; scoring such traces by their final answer instead changes no paired difference in \cref{tab:budget} by more than \budEdgeMax{} problems.

\section{Additional tables}
\label{app:tables}

\begin{table}[H]
\centering
\caption{\capt{Accuracy (\%) against thinking budget, from readouts (one trace per problem).} $\Delta$ is \ours{} minus \base{} in problems with a 95\% paired interval; the last column continues truncated traces to 80K tokens.}
\label{tab:budget}
\scriptsize
\setlength{\tabcolsep}{2.4pt}
\begin{tabular*}{\linewidth}{@{\extracolsep{\fill}}llccccccc@{}}
\toprule
\hd{Set} & \hd{Model} & 4K & 8K & 12K & 16K & 24K & 32K & Uncapped \\
\midrule
Hard set & Qwen3.5-9B & 56.7 & 61.7 & 65.3 & 67.6 & 70.5 & 73.2 & 75.7 \\
Hard set & Concise-9B & 55.6 & 63.8 & 67.5 & 69.6 & 71.3 & 72.9 & 73.7 \\
 & $\Delta$ (problems) & \dci{-14}{-41}{+13} & \dci{+27}{+1}{+53} & \dci{+28}{+3}{+53} & \dci{+25}{0}{+50} & \dci{+11}{-14}{+36} & \dci{-4}{-28}{+20} & \dci{-25}{-49}{-1} \\
\addlinespace[3pt]
Confirmation set & Qwen3.5-9B & 77.8 & 82.6 & 83.4 & 84.4 & 85.4 & 85.5 & 85.7 \\
Confirmation set & Concise-9B & 78.1 & 83.6 & 84.4 & 85.3 & 85.6 & 85.5 & 85.6 \\
 & $\Delta$ (problems) & \dci{+7}{-23}{+37} & \dci{+22}{-1}{+46} & \dci{+23}{0}{+46} & \dci{+21}{0}{+42} & \dci{+4}{-17}{+25} & \dci{0}{-20}{+20} & \dci{-2}{-22}{+18} \\
\bottomrule
\end{tabular*}
\end{table}

\begin{table}[H]
\centering
\caption{\capt{Accuracy (\%) by thinking length, with the number of traces in parentheses (one trace per problem at a 32K budget; ``cap'' are traces that reached the budget).}}
\label{tab:bins}
\scriptsize
\setlength{\tabcolsep}{2.6pt}
\begin{tabular*}{\linewidth}{@{\extracolsep{\fill}}llcccccc@{}}
\toprule
\hd{Benchmark} & \hd{Model} & $<$2K & 2--4K & 4--8K & 8--16K & 16--32K & cap \\
\midrule
Competition mathematics & Qwen3.5-9B & -- & -- & {\color{ohmuted}(6)} & 100 (42) & 91 (107) & 37 (124) \\
 & Concise-9B & -- & {\color{ohmuted}(4)} & 94 (36) & 96 (69) & 77 (100) & 17 (70) \\
\addlinespace[1pt]
Hard MATH & Qwen3.5-9B & {\color{ohmuted}(16)} & 99 (70) & 98 (110) & 99 (117) & 100 (66) & {\color{ohmuted}(18)} \\
 & Concise-9B & 98 (41) & 100 (122) & 97 (136) & 99 (73) & {\color{ohmuted}(18)} & {\color{ohmuted}(7)} \\
\addlinespace[1pt]
SuperGPQA & Qwen3.5-9B & {\color{ohmuted}(13)} & 55 (38) & 59 (143) & 61 (188) & 65 (147) & 41 (71) \\
 & Concise-9B & 52 (25) & 62 (152) & 60 (184) & 61 (152) & 51 (71) & {\color{ohmuted}(16)} \\
\addlinespace[1pt]
MATH test & Qwen3.5-9B & {\color{ohmuted}(14)} & 98 (63) & 97 (120) & 99 (162) & 99 (101) & {\color{ohmuted}(16)} \\
 & Concise-9B & 100 (41) & 99 (147) & 100 (157) & 97 (108) & 91 (23) & -- \\
\addlinespace[1pt]
Science & Qwen3.5-9B & {\color{ohmuted}(8)} & 86 (28) & 72 (129) & 65 (190) & 66 (165) & 51 (80) \\
 & Concise-9B & 83 (24) & 74 (73) & 67 (217) & 67 (189) & 58 (77) & 30 (20) \\
\addlinespace[1pt]
Logical constraints & Qwen3.5-9B & 98 (41) & 91 (134) & 86 (229) & 84 (110) & 89 (57) & {\color{ohmuted}(4)} \\
 & Concise-9B & 93 (55) & 89 (244) & 82 (170) & 93 (85) & {\color{ohmuted}(17)} & {\color{ohmuted}(4)} \\
\addlinespace[1pt]
MMLU-Pro (subset) & Qwen3.5-9B & 100 (21) & 92 (40) & 89 (85) & 84 (87) & 62 (39) & {\color{ohmuted}(17)} \\
 & Concise-9B & 97 (35) & 92 (89) & 85 (102) & 71 (42) & {\color{ohmuted}(19)} & {\color{ohmuted}(2)} \\
\addlinespace[1pt]
MMLU-Pro & Qwen3.5-9B & 97 (462) & 91 (932) & 87 (1,690) & 80 (1,452) & 72 (682) & 44 (214) \\
 & Concise-9B & 96 (771) & 89 (1,792) & 80 (1,820) & 71 (760) & 53 (257) & 44 (32) \\
\bottomrule
\end{tabular*}
\end{table}

\begin{table}[H]
\centering
\caption{\capt{Explicit answer statements in the thinking.} ``Stated early'' is the share of traces with a stable explicit statement of the final answer before the end; ``Slack'' is the share of thinking tokens after it, next to the readout-based share of \cref{tab:slack}; the last column gives the share of readouts that equal the final answer before / after the statement.}
\label{tab:text}
\footnotesize
\begin{tabular*}{\linewidth}{@{\extracolsep{\fill}}llrrrr@{}}
\toprule
\hd{Set} & \hd{Model} & Stated early (\%) & Slack, text (\%) & Slack, readouts (\%) & Readouts agreeing (\%) \\
\midrule
Hard set & Qwen3.5-9B & 50 & 8.1 & 54.0 & 76 / 95 \\
Hard set & Concise-9B & 30 & 4.5 & 41.0 & 65 / 93 \\
Confirmation set & Qwen3.5-9B & 59 & 10.3 & 63.8 & 86 / 98 \\
Confirmation set & Concise-9B & 39 & 6.0 & 48.3 & 77 / 100 \\
MMLU-Pro & Qwen3.5-9B & 61 & 11.2 & -- & -- \\
MMLU-Pro & Concise-9B & 44 & 6.5 & 48.2 & 80 / 96 \\
Development suite & Qwen3.5-9B & 57 & 8.8 & -- & -- \\
Development suite & Concise-9B & 36 & 4.5 & -- & -- \\
\bottomrule
\end{tabular*}
\end{table}

\begin{table}[H]
\centering
\caption{\capt{Calibration of the pace comparison (hard set).} Predicted correct counts use \base{}'s readouts at each trace's stopping length; the three further \base{} traces are its 64K traces scored at 32K (a few problems without a readout at 32,768 tokens are left out). The residual is actual minus predicted, with a 95\% paired interval.}
\label{tab:pace}
\footnotesize
\begin{tabular*}{\linewidth}{@{\extracolsep{\fill}}lrrrl@{}}
\toprule
\hd{Traces} & $n$ & \hd{Actual} & \hd{Predicted} & Residual [95\% CI] \\
\midrule
Qwen3.5-9B, further sample 1 & 1,266 & 922 & 905 & \dci{+17}{-6}{+40} \\
Qwen3.5-9B, further sample 2 & 1,264 & 929 & 897 & \dci{+32}{+8}{+57} \\
Qwen3.5-9B, further sample 3 & 1,270 & 912 & 903 & \dci{+9}{-16}{+34} \\
\textbf{Concise-9B} & 1,276 & 930 & 866 & \dci{+64}{+36}{+92} \\
\bottomrule
\end{tabular*}
\end{table}

\begin{table}[H]
\centering
\caption{\capt{Four traces per model and problem (hard set).} \base{}: its native trace continued to 80K where truncated, plus three further traces capped at 64K (\fourHardBaseStill{} of them still truncated over the hard set, \fourSgpqaBaseStill{} on SuperGPQA); \ours{}: its native trace and three 32K traces, each continued to 80K where truncated. Escalation as in \cref{tab:dev}. The SuperGPQA comparison of escalations was pre-registered with a $\pm$2-point (12-problem) margin and a token-ratio limit of 0.60; the other rows are descriptive. All \fourHardN{} hard-set problems are included, as registered, including the \ovHard{} excluded elsewhere for overlap with training.}
\label{tab:four}
\footnotesize
\begin{tabular*}{\linewidth}{@{\extracolsep{\fill}}lrrllr@{}}
\toprule
\hd{Comparison} & \hd{Qwen} & \hd{Concise} & $\Delta$ [95\% CI] & 90\% CI & \hd{Ratio} \\
\midrule
\grouprow{6}{SuperGPQA (600)}
\quad Escalation vs.\ escalation & 370 & 363 & \dci{-7}{-23}{+9} & \ensuremath{[-20,\ +6]} & 0.61\x{} \\
\quad Escalation vs.\ majority of 4 & 372 & 363 & \dci{-9}{-24}{+7} & \ensuremath{[-22,\ +4]} & 0.39\x{} \\
\quad Single trace vs.\ single trace & 359 & 354 & \dci{-5}{-25}{+14} & \ensuremath{[-21,\ +11]} & 0.60\x{} \\
\grouprow{6}{Mathematics (679)}
\quad Escalation vs.\ escalation & 624 & 611 & \dci{-13}{-24}{-2} & \ensuremath{[-22,\ -4]} & 0.69\x{} \\
\quad Escalation vs.\ majority of 4 & 624 & 611 & \dci{-13}{-24}{-2} & \ensuremath{[-23,\ -3]} & 0.44\x{} \\
\quad Single trace vs.\ single trace & 610 & 590 & \dci{-20}{-34}{-6} & \ensuremath{[-32,\ -8]} & 0.67\x{} \\
\grouprow{6}{Whole hard set (1,279)}
\quad Escalation vs.\ escalation & 994 & 974 & \dci{-20}{-39}{-1} & \ensuremath{[-36,\ -4]} & 0.65\x{} \\
\quad Escalation vs.\ majority of 4 & 996 & 974 & \dci{-22}{-41}{-3} & \ensuremath{[-38,\ -6]} & 0.42\x{} \\
\quad Single trace vs.\ single trace & 969 & 944 & \dci{-25}{-49}{-1} & \ensuremath{[-45,\ -5]} & 0.64\x{} \\
\bottomrule
\end{tabular*}
\end{table}

\end{document}